\documentclass[sigconf]{acmart}
\usepackage{multicol}
\usepackage{multirow}
\usepackage{stfloats}
\usepackage{float}
\usepackage{epstopdf}

\AtBeginDocument{%
  \providecommand\BibTeX{{%
    \normalfont B\kern-0.5em{\scshape i\kern-0.25em b}\kern-0.8em\TeX}}}

\copyrightyear{2023}
\acmYear{2023}
\setcopyright{acmlicensed}\acmConference[WWW '23]{Proceedings of the ACM
Web Conference 2023}{May 1--5, 2023}{Austin, TX, USA}
\acmBooktitle{Proceedings of the ACM Web Conference 2023 (WWW '23), May
1--5, 2023, Austin, TX, USA}
\acmPrice{15.00}
\acmDOI{10.1145/3543507.3583314}
\acmISBN{978-1-4503-9416-1/23/04}
\begin{document}

\title{Improving (Dis)agreement Detection with Inductive Social Relation Information From Comment-Reply Interactions}


\author{Yun Luo}
\affiliation{%
  \institution{School of Engineering, Westlake University}
  \city{Hangzhou}
  \country{China}}
\email{luoyun@westlake.edu.cn}

\author{Zihan Liu}
\affiliation{%
  \institution{AI Lab, Research Center for Industries of the Future, Westlake University}
  \city{Hangzhou}
  \country{China}}
\email{liuzihan@westlake.edu.cn}

\author{Stan Z. Li}
\affiliation{%
  \institution{AI Lab, Research Center for Industries of the Future, Westlake University}
  \city{Hangzhou}
  \country{China}}
\email{stanzq@westlake.edu.cn}

\author{Yue Zhang$^*$}
\affiliation{%
  \institution{School of Engineering, Westlake University}
  \authornote{Corresponding author}
  \city{Hangzhou}
  \country{China}}
\email{zhangyue@westlake.edu.cn}







\renewcommand{\shortauthors}{}

\begin{abstract}
(Dis)agreement detection aims to identify the authors' attitudes or positions (\textit{{agree, disagree, neutral}}) towards a specific text.  It is limited for existing methods merely using textual information for identifying (dis)agreements, especially for cross-domain settings. Social relation information can play an assistant role in the (dis)agreement task besides textual information. We propose a novel method to extract such relation information from (dis)agreement data into an inductive social relation graph, merely using the comment-reply pairs without any additional platform-specific information. The inductive social relation globally considers the historical discussion and the relation between authors. Textual information based on a pre-trained language model and social relation information encoded by pre-trained RGCN are jointly considered for (dis)agreement detection. Experimental results show that our model achieves the state-of-the-art performance for both the in-domain and cross-domain tasks on the benchmark -- DEBAGREEMENT. We find social relations can boost the performance of the (dis)agreement detection model, especially for the long-token comment-reply pairs, demonstrating the effectiveness of the social relation graph. We also explore the effect of the knowledge graph embedding methods, the information fusing method, and the time interval in constructing the social relation graph, which shows the effectiveness of our model.
\end{abstract}





\maketitle

\begin{figure}
    \centering
    \includegraphics[width=0.95\hsize]{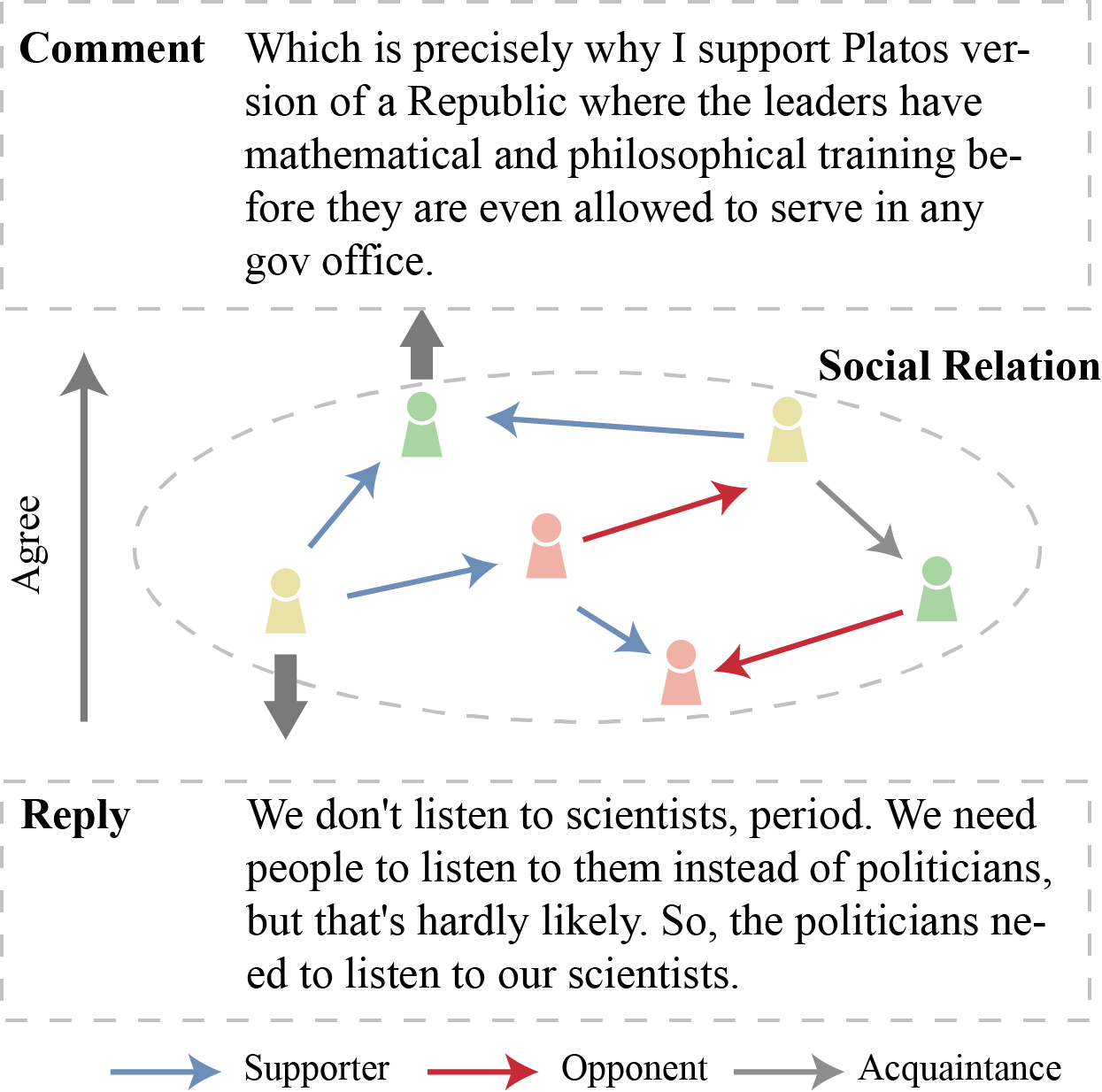}
    \caption{Examples for (dis)agreement detection in the DEBAGREEMENT dataset. }
    \label{example}
    \vspace{-0.5cm}
\end{figure}

\section{Introduction}
Automatic elicitation of semantic information has attracted increasing attention in recent years to the widespread social platforms online. The tasks contain sentiment analysis, sarcasm detection, stance detection, etc. We focus on the task of (dis)agreement detection, which aims to identify the authors' attitudes or positions (\textit{{agree, disagree, neutral}}) towards a specific text  \cite{pougue2021debagreement}. This task falls under the field of stance detection and opinion mining. 
For example, to the text `\textit{Peace is sometimes a translation of Shalom, which also carries the meaning of wellbeing. It speaks to the heart of what Peace is about.}', the reply
`\textit{Someone explains to me how climate activism relates to peace I feel like it's a bit unrelated}' expresses a disagreeing stance. The task of (dis)agreement detection is crucial in understanding the societal polarisation and spread of ideas online \cite{ribeiro2017everything,tan2016winning,rosenthal2015couldn}.

There are several challenges for (dis)agreement detection. One salient challenge is that the textual information is limited for the task \cite{pougue2021debagreement}, and when a human detects the (dis)agreement of a reply to a specific comment, some commonsense knowledge or contextual understanding assists in the identification. Taking the first example in Figure \ref{example}, the comment talks about leaders' mathematical and philosophical training. However, the reply is about the importance of scientists, not politicians, which has different textual features. It is difficult for models to correctly identify the (dis)agreement solely based on the textual features. In addition, it remains challenging to detect (dis)agreements in cross-domain settings \cite{pougue2021debagreement,aldayel2021stance}, where the topics or contents of testing data are different from the training data. The language expressions for stance-related text usually vary across different topics.
Suppose the (dis)agreement detection model is trained on the data of Republican, like the example in Figure 1, but tested on the topic of Climate. In that case, models can be confused in giving identification.

It has been identified in sociology and psychology studies that individuals' opinions are significantly affected by social relations and online contents \cite{cheng2020dynamic,lorenz2020behavioural,jiao2021active}. Accordingly, some studies use contextual information from Twitter, such as hashtags or retweets, to solve the challenges mentioned above \cite{dey2017assessing,samih2021few}. For example, \citet{dey2017assessing} propose a latent concept space to obtain the stance similarity 
using twitter hashtags for identifying stance. \citet{samih2021few} propose a classification method based on the accounts he/she retweeted, computing its
similarity to all users in the training set.   Nevertheless, the features limit the extensibility of the models, which most datasets or platforms lack. In addition to this specific information, individuals' relations can also be reflected by the interactions between them, which is common, and easily obtained in most scenarios. However, relatively little work applies the information due to the lack of suitable datasets.

Recently, \citet{pougue2021debagreement} propose a large dataset in real-world online discussions (42,804 comment-reply pairs) on Reddit\footnote{reddit.com: the 20th most visited site globally as of March 2020}, which contains information of authors and the temporal order (common information on most social platforms). The dataset provides a testbed for investigating general social information's effect and how it enhances (dis)agreement detection models.   
In particular, the dataset contains contextual information (authorship, post, timestamp, etc.) and comment-reply pairs for (dis)agreement detection. We reform the comment-reply pairs to a social relation graph to detect (dis)agreement with social information, which facilitates the (dis)agreement detection.
 For the examples in Figure 1, if social relation information is effectively used so that the model knows that the authors of the replies are supporters of comment authors (obtained from previous interactions), it becomes simple to identify the (dis)agreement of the comment-reply pairs. In addition, for the cross-domain setting, the model can be more accurate with the use of the social relation information, where implicit relations such as `A friend of my enemy is my enemy' \cite{cartwright1956structural} can also be effectively used by message passing from graph neural networks.

Individuals tend to maintain their initial beliefs even in the face of evidence that contradicts them, which is called belief perseverance in psychology \citep{anderson1992belief,guenther2008self}. Thus, individuals tend to insist on their (dis)agreement with others on a specific issue. Inspired by the effectiveness of graph neural networks in extracting the representation of structured data \cite{wu2020comprehensive,liu2022towards,jin2020graph}, we propose a novel method to extract social relations from temporal comment-reply interactions to an inductive social relation graph, which gives general information on different social platforms and offline scenarios. 
We pre-train a graph autoencoder to encode social relation information through a relational graph network (RGCN) \cite{Schlichtkrull2018ModelingRD} encoder and a knowledge graph embedding (KGE) decoder DistMult \cite{yang2014embedding}. 
The social relations information encoded by the graph autoencoder is fused with textual information from pre-trained language models such as BERT \cite{devlin2018bert}, and RoBERTa \cite{liu2019roberta}  to identify the (dis)agreement.

Experiments show that our model achieves state-of-the-art performance in in- and cross-domain settings for (dis)agreement detection on the standard benchmark \cite{pougue2021debagreement}. We prove the effectiveness of social relation features on BiLSTM, BERT, and RoBERTa for (dis)agreement detection. Then we demonstrate that the model performs better for long-token comment-reply pairs. We also show the significance of each module in our model, such as the reconstruction loss of graph features and the pre-training of the graph autoencoder. To the best of our knowledge, we are the first to consider the general inductive social relation information from comment-reply pairs for (dis)agreement detection.
The codes and trained models can be found at {\url{https://github.com/LuoXiaoHeics/StanceRel}}.

The contributions of our paper can be summarized as follows:
\begin{enumerate}
    \item We propose a novel method to extract relation information from (dis)agreement data into an inductive social relation graph, merely using the comment-reply pairs without any additional platform-specific information. 
    \item We propose a (dis)agreement detection model jointly considering the textual information from pre-trained language models and social relation information from pre-trained RGCN.
    \item Experimental results show that our model achieves state-of-the-art performance for both the in-domain and cross-domain tasks. We also show the effectiveness of our models through various analyses.
\end{enumerate}

\section{Related Work}
(Dis)agreement detection is a sub-task of stance detection \cite{pougue2021debagreement,li2021p,luo2022exploiting}, (also known as stance classification \cite{walker}, stance identification \cite{zhang2017we}, stance prediction \cite{qiu2015modeling}, debate-side classification \cite{anand2011cats}, and debate stance classification \cite{hasan-ng-2013-stance}). Many models are proposed to solve the task of stance detection or (dis)agreement detection by solely using textual information, while some studies have used graph (or network) features to boost the performance of stance detection or (dis)agreement detection, such as interaction networks, preference networks, and connection networks. \citet{borge2015content,darwish2018predicting} and \citet{darwish2020unsupervised} propose to use the relations of retweet data. \citet{dey2017assessing} and \citet{samih2021few} make use of hashtags to infer Twitter users' stances. 

Existing work also considers incorporating social context \cite{keskar2019ctrl} and structured knowledge \cite{colon2021combining} into language models to boost the performance on natural language processing tasks. However, previous datasets on stance detection mostly merely provide textual information. Some work that uses graph features is specific to a Twitter discussion, using the hastags or retweets, which limits the extensibility of the model. Unlike previous studies, we propose a simple method to construct the social relation graph using the (dis)agreement data with authors and temporal orders, which are common information in most social platforms or debate situations, boosting the extensibility of using social relation graph for disagreement detection.

We use knowledge graph embedding (KGE) methods to pre-train the node embeddings of the authors, which are widely used for encoding knowledge graph concepts. KGE methods can effectively simplify the manipulation while preserving the knowledge graph's inherent structure and achieving remarkable performance in the downstream tasks such as knowledge graph completion, and relation extraction  \cite{nickel2016holographic,yang2014embedding,bordes2013translating}. Prior work can be divided into translational distance models using distance-based scoring functions and semantic matching models using similarity-based ones \cite{Wang2017}. In this paper, we use the idea of knowledge graph embedding to pre-train a graph autoencoder to extract social information, assisting (dis)agreement detection.

\section{Method}
The architecture of our model is illustrated in Figure \ref{fig1}, which contains two components: (1) relation graph encoding, which extracts social relation information (Section 3.2); (2) (dis)agreement detection with relation information (Section 3.3). 

\subsection{Task Description}
We formulate the (dis)agreement detection task as a classification task. Formally, let $D= \{c_i, t_i, y_i, n^c_i, n^t_i\}_{i=1}^N$ be a dataset with N examples, each consisting of a comment $c_i$ from author $n^c_i$, a reply $r_i$ from author $n^t_i$, and a stance label $y_i$ from $r_i$ to $n^c_i$ through the comment-reply pair. The task is to predict a stance label $\hat{y}\in \{agree, disagree, neutral\}$ for each comment-pair, based on the definition of \citet{pougue2021debagreement}.

\subsection{Relation Graph Autoencoder}
We denote the relation graph as a directed graph $\mathcal{G = \{N,E,R}\}$, with nodes  (authors) $n_i \in \mathcal{N}$ and labeled edges $(n_i,r,n_j) \in \mathcal{E}$, where $r \in \mathcal{R}$ is the relation type of the edge from $n_i$ to $n_j$. The relation types include \{\textit{supporter, opponent}, \textit{acquaintance}, \textit{interaction}\}.

\begin{figure}
    \centering
    \includegraphics[width=0.85\hsize]{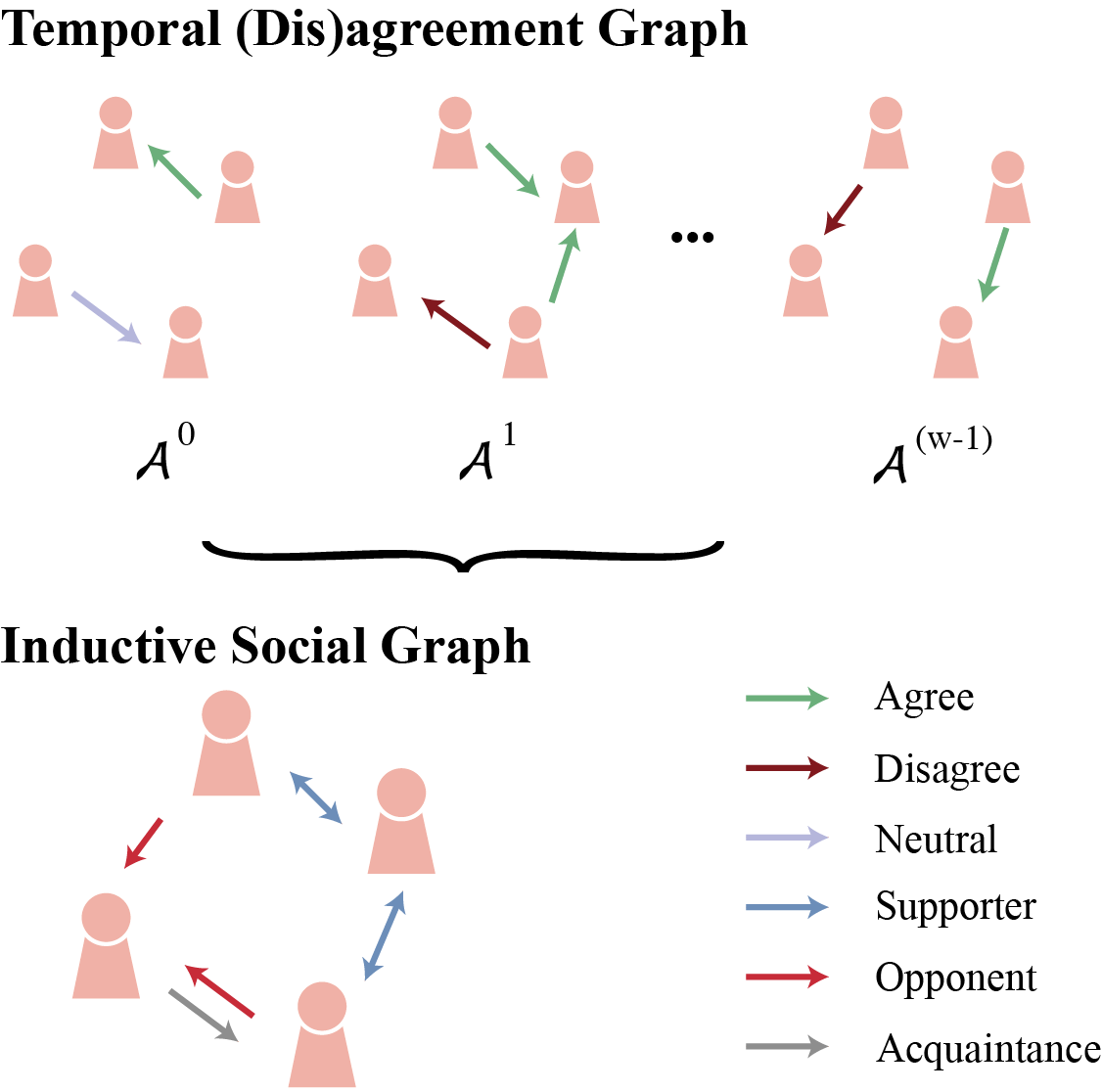}
    \caption{The illustration of the construction of the social relation graph using the temporal order information. }
    \label{temporal}
    \vspace{-5mm}
\end{figure}

\begin{figure*}
    \centering
    \includegraphics[width=0.9\hsize]{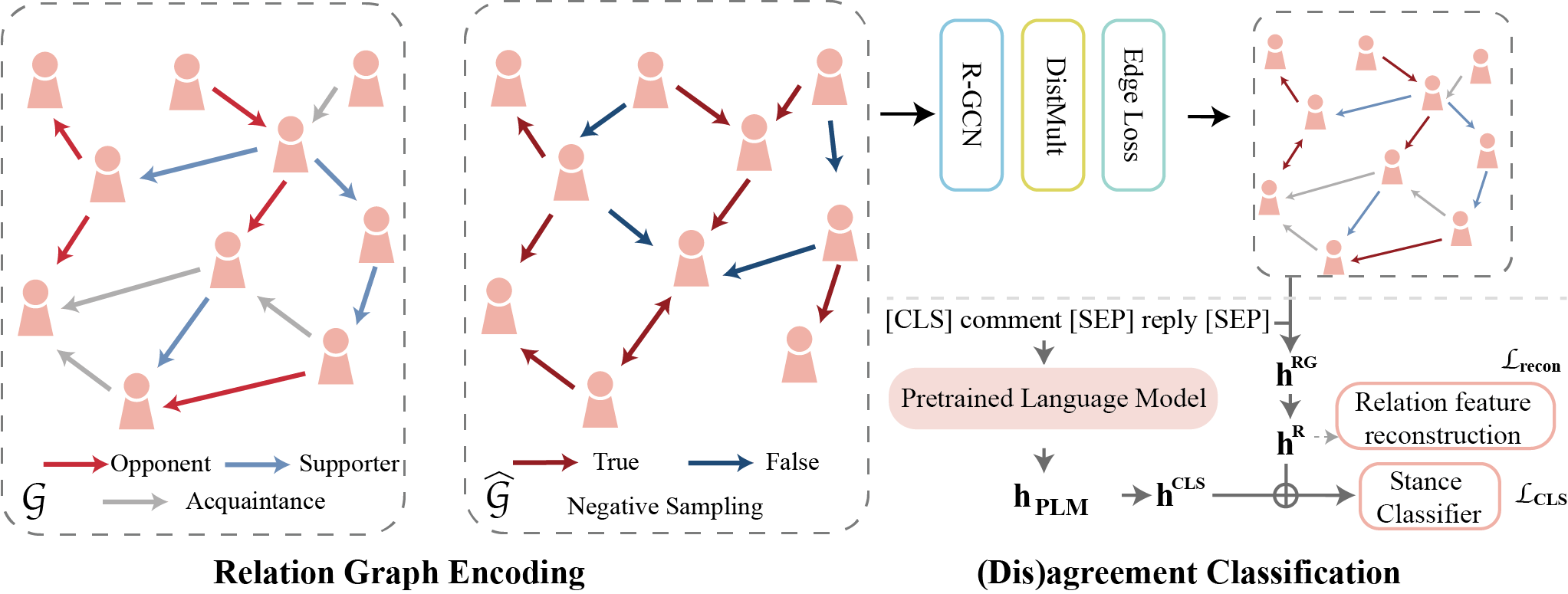}
    \caption{Framework of our proposed model, which contains two components, (1) relation graph encoding, (2) (dis)agreement detection with social relation information.}
    \label{fig1}
    \vspace{-4mm}
\end{figure*}

\textbf{Social Relation Graph Construction.} To construct the relation graph, we first extract the set of all authors in the dataset, corresponding to the node set $\mathcal{N}$. The time interval to aggregate the social relations is a significant factor due to the temporal effects of social relations between individuals. Inspired by \cite{karsai2011small,luo2021opinion}, we model the temporal network by weighting the links with frequencies to obtain the type of social relation. For the sequence of the graph weighted adjacent matrix (snapshots) $S(w,\tau) = [\mathcal{A}^0,\mathcal{A}^1,...,\mathcal{A}^{(w-1)}]$ ($\mathcal{A}^k$ is the graph weighted adjacent matrix during time period $[k\tau,(k+1)\tau]$), the inductive graph is drawn from the interactions that appear during the timescale $w\tau$ (Figure \ref{temporal}). For each graph weighted adjacent matrix $\mathcal{A}^k$, if the author $n_i$ expresses an agree/disagree/neutral stance towards $n_j$ in a comment-reply pair, the value $a^k_{i,j} = +1/-1/0$, and if there are multiple interactions between them, the most frequent opinion  (agree/disagree/neutral) are considered to determine $a^k_{i,j}$.  Then the weighted adjacent matrix of the inductive graph is $\mathcal{A}^* = \mathcal{A}^0+\mathcal{A}^1+...+\mathcal{A}^{(w-1)}$. The triplet $(n_i,r,n_j)$ is in  relation graph $\mathcal{G}_R$ as follows:
\begin{displaymath}
  r = \left\{ \begin{array}{lll}
  supporter & \textrm{if $a^*_{ij}>0$},\\
  opponent & \textrm{if $a^*_{ij}<0$},\\
  acquaintance & \textrm{if $a^*_{ij}=0$ and $a^k_{ij}\ne 0$,} \\ 
  \end{array} \right.
  \end{displaymath}
  and $(n_i,r,n_j)$ is not in $\mathcal{G}_R$ in other situations. 
  
 In order to avoid label leaking in development and test sets, we add another type of relation \textit{interaction} for the edges unseen in the training set but appear in the development and test sets. The node feature can be normally observed in the semi-supervised learning on graph neural network tasks \cite{chong2020graph,van2020survey}. It also aims to solve the issue that node features would be unknown if the nodes are not added to the social relation graph before pre-training. To avoid the over-fitting of training the model, we randomly select edges in $\mathcal{G}_R$ with probability $\rho$ to be \textit{interaction} edges.

To obtain the social relation information, a graph autoencoder is adopted following  \citet{Schlichtkrull2018ModelingRD}. An incomplete set (randomly sampled with 50\% probability) of edges $\hat{\mathcal{E}}$ from $\mathcal{E}$ in $\mathcal{G}_R$ is fed as the graph autoencoder input. The incomplete set $\hat{\mathcal{E}}$ is negatively sampled to a complete set of samples denoted $\mathcal{U}$ (details in Training). Then we assign the possible edges $(n_i,r,n_j) \in \mathcal{U}$  with scores, which are used to determine the probability of whether the edges are true in $\mathcal{E}$. Relational author network (RGCN) \cite{Schlichtkrull2018ModelingRD} is applied to the encoder to obtain the latent feature representations of authors, and a DistMult scoring decoder \cite{yang2014embedding} is used to recover the missing edges. 

\textbf{Encoder}.
 The RGCN  module serves to accumulate relational evidence in multiple inference steps. In each step, a neighborhood-based convolutional feature transformation process uses the related authors to induce an enriched author-aggregated feature vector for each author. Two stacked RGCN encoders are applied to encode the social information. The parameters of author feature vectors are initialized with $\textbf{u}_i$. Then the vectors are transformed into relation-aggregated feature vectors $\textbf{h}_i \in \mathbb{R}^d$ using the RGCN encoders:
\begin{equation}
    {f}(x_i,l) = \sigma(W_0^{(l)}x_i+\sum_{r\in\mathcal{R}}\sum_{j\in N^r_i} \frac{1}{n_{i,r}}W_r^{(l)}x_j),
    \nonumber 
\end{equation}
\vspace{-2mm}
\begin{equation}
    \textbf{h}_i = \textbf{h}_i^{(2)} = f(\textbf{h}_i^{(1)},2) \ ; \ \textbf{h}_i^{(1)} = f(\textbf{u}_i,1),
\end{equation}
where $f$ is the encoder network (requiring inputs of feature vector $x_i$ and the rank of the layer $l$); $N^r_i$ is the neighbouring authors $i$ with the relation $r\in \mathcal{R}$; $n_{i,r}$ is a normalization constant, set in advance $n_{i,r} = |N^r_i|$ or learned by network learning; $\sigma$ is the activation function such as ReLU, and $W_r^{(1)},W_0^{(1)},W_r^{(2)},W_0^{(2)}$ are learnable parameters though training.

\textbf{Training}.  We use DistMult factorization as the decoder to assign scores. For a given triplet $(n_i,r,n_j) $, the score can be obtain as follows:
\begin{equation}
    s(n_i,r,n_j) = \sigma(\textbf{h}_{n_i}^TR_r\textbf{h}_{n_j}),
\end{equation}
where $\sigma$ is a logistic function; $\textbf{h}_{n_i}, \textbf{h}_{n_j} \in \mathbb{R}^n$ are the encoding feature vectors through the graph encoder for author $n_i$ and $n_j$; every type of relation $r\in \mathcal{R}$ is associated with a diagonal matrix $R_r \in \mathbb{R}^{n\times n}$.

The method of negative sampling \cite{Schlichtkrull2018ModelingRD} is used for training our graph autoencoder module. First, we randomly corrupt the true triplets, i.e., triplets in $\hat{\mathcal{E}}$, to create an equal number of false samples. We corrupt the triplets by randomly modifying the connected authors or relations, creating the overall set of samples $\mathcal{U}$. The training objective is a binary classification between true/false (denoted as $y$) triplets  with a cross-entropy loss function:
\begin{equation}
\begin{aligned}
\mathcal{L}_{\mathcal{G}'} = & -\frac{1}{2|\hat{\mathcal{E}'}|}\sum_{(n_i,r,n_j,y)\in \mathcal{U}}(y\,log\ s(n_i,r,n_j) \\
&+(1-y)log(1-s(n_i,r,n_j))).
\end{aligned}
\end{equation}
\subsection{(Dis)agreement Detection}
\textbf{Relation Feature Encoding}.
After training the graph autoencoder, in order to extract the author-specific relation graph feature for the comment $c_i$ and the reply $t_i$, we denote $n^i_c$ and $n^i_t$ as the authors for $c_i$ and $t_i$, respectively. Then we extract a sub-graph $\mathcal{G}_A$ from $\mathcal{G}_R$, which contains all the authors on the graph within the vicinity of radius 1 from $n^i_c$ and $n^i_t$. Next, we make a forward pass of $\mathcal{G}_A$ through the encoder of graph autoencoder to obtain the feature vectors $\textbf{h}_j$ for all unique authors $j$ in $\mathcal{G}_A$. The average of  feature vectors $\textbf{h}_j$ for all unique authors in $\mathcal{G}_A$ is regarded as the relation graph feature vector $\textbf{h}^{RG}$:
\begin{equation}
\textbf{h}^{RG} = RGCN(\mathcal{G}_A).
\end{equation}

The relation graph feature vector $\textbf{h}^{RG}$ is fed into a linear layer to obtain hidden states $\textbf{h}^R$:
\begin{equation}
    \textbf{h}^R = W_R\;\textbf{h}^{RG}+b_R
\end{equation}
where $W_R$ and $b_R$ are the trained parameters of the linear layer.

\textbf{Textual Feature Encoding}. Pre-trained language models (PLMs), such as BERT\cite{devlin2018bert}, RoBERTa \cite{liu2019roberta}, and GPT3 \cite{radford2019language}, have been proven effective in various NLP applications, which are pre-trained on the large-scale unlabelled corpus. Taking BERT, for example, it uses a bidirectional transformer on single or multiple sentences. We take $[CLS]\ c_i \ [SEP] $ $\ t_i \ [SEP]$ as the input $x_i$ for our model, where $[CLS]$ refers to the first token of the sequence and $[SEP]$ is used to separate sequences. The input $x_i$ is fed into PLMs such as BERT and RoBERTa to obtain its hidden states:
\begin{equation}
         {\textbf{h}_{PLM}} = PLM(x_i).
\end{equation}
The hidden state of $[CLS]$ token is adopted as the representation of the comment-reply pairs.

\textbf{(Dis)agreement Classification.} 
 The hidden states vectors of $\textbf{h}^R$ and  $\textbf{h}^{CLS}$ are concatenated for classification:
\begin{equation}
    p = Softmax(W[\textbf{h}^{CLS},\textbf{h}^{R}]+b),
\end{equation}
where $W$ and $b$ are the parameters and $p$ is the probability distribution on the three (dis)agreement labels. 

\textbf{Training.} The training loss $\mathcal{L}_{train}$ consists of a classification term and a reconstruction term, denoted as:
\begin{equation}
    \mathcal{L}_{train} = \mathcal{L}_{stance}+\mathcal{L}_{recon}.
\end{equation}
Given the input and its golden label $(x_i,y_i)$, the $\mathcal{L}_{stance}$ for classifying (dis)agreement is a cross-entropy loss:
\begin{equation}
    \mathcal{L}_{stance} = -\frac{1}{|N|}\sum_{(x_i,y_i)}y_i\,log\ p(y_i),
\end{equation}
where $|N|$ is the number of data samples.
To further ensure stronger author invariance constraints of $\textbf{h}_{RG}$, we add a shared decoder layer $D_{recon}$ with a reconstruction loss:
\begin{equation}
    \mathcal{L}_{recon} = -E_{\textbf{h}^{RG}}(||D_{recon}({\textbf{h}^R}) - \textbf{h}^{RG}||_2^2).
\end{equation}

\begin{table}[]

\begin{center}

\begin{tabular}{lccccccc}
\hline

      & r/Br & r/Cl  & r/BLM & r/Re &r/De \\
      \hline \hline
\#nodes & 722                   & 4,580     &2,516           & 8,832        &6,925               \\
\#edges   & 15,745          &5,773           & 1,929             & 9,823         & 9,624          \\
Agree  & 29\%                       & 32\%        & 45\%         & 34\%&42\%      \\
Neutral & 29\%  &28\% &22\% &25\%&22\% \\
Disagree &42\%  &40\% &33\% &41\%&36\%\\
\hline
\end{tabular}
\caption{Statistics on DEBAGREEMENT. Br for the subreddit Brexit; Cl for the subreddit Climate; BLM for the subreddit BLM; Re for the subreddit Republican and De for the subreddit Democrats, henceforth. }
      \label{table1}
      \vspace{-8mm}
\end{center}

\end{table}

\begin{table}[]

\begin{center}

\begin{tabular}{lcccccccc}
\hline

      & r/Br & r/Cl  & r/BLM & r/Re &r/De &All \\
      \hline \hline
\#Supporter & 2,159       &989&511&1,882&2,299&7,833 \\
\#Opponent   &3,040   &1,304&357&2,170&1,957&8,820  \\
\#Interaction  & 7,613 &3,383&1,039&5,723&5,276&23,004  \\
Degree  &35.39&2.48 &1.51  &2.22&2.75&3.43 \\
Betweenness  &1.54&0.49&0.01&0.22&0.52&0.53 \\
\hline
\end{tabular}
\caption{Statistics metrics on the inductive social relation graph and the subgraph of each subreddit. Degree and betweenness are the averaged metrics on each subgraph, which indicate the graph centrality. }
      \label{graph}
      \vspace{-8mm}
\end{center}

\end{table}

\begin{table*}[]

\begin{center}
 \setlength{\tabcolsep}{2.3mm}{\begin{tabular}{lccccccccccc}
\hline

\multirow{2}{*}{Model}            & \multicolumn{3}{c}{Agree}                                                                                                     & \multicolumn{3}{c}{Disagree}                                                                                                      & \multicolumn{3}{c}{Neutral} &\multicolumn{2}{c}{All}                                                                                                  \\ \cline{2-12}  & 
Prec& Rec                      & F1                                                              & Prec                               & Rec                      & F1                                                     & Prec                               & Rec                      & F1  & Acc &M-F1  \\ \hline \hline

BiLSTM &47.29 &47.85&47.56&47.86&61.96&54.00&44.44&25.87&32.70&47.11&44.75\\

BERT-sep & 68.92 & 68.26 &68.44  & 68.79 &{73.29} &70.58 &53.29 &48.55 & 50.80 & 64.68&63.27 \\
BERT-joint & 67.88 &{67.78}&66.30 & 68.84&74.80&70.36& 54.44&48.12&50.28 &65.50& 63.59\\
RoBERTa-joint &\textbf{72.28}&69.18&70.56&74.11&69.80&71.89 &51.31&58.67&54.57&66.78&65.67 \\
\hline
\textbf{Ours} \\ 
BiLSTM-rel&50.35& 57.65&53.75 &51.87&55.71&53.77&42.23 & 28.79 &34.17&49.62&47.23\\
BERT-rel  &70.15&70.60&70.35&73.62&71.19&72.34&52.52&54.68&53.51&66.82&65.40\\
RoBERTa-rel &{70.97}&\textbf{72.01}&\textbf{71.44}&\textbf{75.62}&\textbf{73.01}&\textbf{74.27}&\textbf{54.16}&\textbf{55.95}&\textbf{55.02}&\textbf{68.38}&\textbf{66.91} \\
\hline 

\end{tabular}
}
\end{center}
\caption{In-domain testing results. The models are trained on the five subreddits and tested on the corresponding test data. (Prec , Rec, F1, Acc and M-F1 for the metrics of precision, recall, micro-F1 score, accuracy and macro-F1 score, henceforth).}
\label{results1}
\vspace{-6mm}
\end{table*}

\section{Experiments}
We verify the effectiveness of social relation information for the in-domain (train the model on all the subreddits and evaluate it on the corresponding test data) in Section 4.3 and cross-domain tasks (train the model on four subreddits and evaluate it on the one subreddit left) in Section 4.4. We also carry out further analysis of our model in Section 4.5.

\subsection{Settings}
\textbf{Dataset}: We adopt the  dataset- DEBAGREEMENT \cite{pougue2021debagreement} for (dis)agree- ment detection. The dataset consists of 42,804 comment-reply pairs from the popular discussion website reddit with authorship and temporal information. The data topics include Brexit, Climate, BlackLivesMatter, Republican, and Democrats. The statistics of the dataset are shown in Table \ref{table1}. As shown in the dataset, the interactions of the dataset are sparse, especially in the subreddits BlackLivesMatter and Republican.

\textbf{Training Details.} We perform experiments using the official pre-trained BERT \cite{devlin2018bert} and RoBERTa \cite{liu2019roberta} models  provided by Huggingface \footnote{https://huggingface.co/}. We train our model on 1 GPU (Nvidia GTX2080Ti) using the Adam optimizer \cite{kingma2014adam}. To construct the relation graph, we use the probability $\rho = 0.3$ to select edges in the training set to be \textit{interaction} edges. We show the statistics of the inductive social relations in Table \ref{graph}. For training the graph autoencoder, the initial learning rate is 1e-2, the epoch is 2e3, the batch size is 1e5, and we take each edge as the temporal 
graph matrix $\mathcal{A}^i$ for the reason that the interactions of authors in the dataset are sparse (23,101 nodes and 42,804 edges). For the (dis)agreement detection training process, the initial learning rate is 1.5e-5, the max sequence length is 256, the batch size for training is 8 for BERT-based/RoBERTa-based models, and the models are trained for three epochs. We split the data into 80\%/10\%/10\% train/val/test sets while maintaining the temporal order, where testing is done on the latest data. We adopt the macro-F1 score to find the best model configuration, and the main results reported are averaged on five different runs.

\textbf{Baselines.} The standard benchmark \cite{pougue2021debagreement} does not contain platform-specific information such as hashtags or retweets,  we provide several baselines for (dis)agreement detection, such as BiLSTM-based models -- BiLSTM, BiLSTM-rel, BERT-based models -- BERT-sep, BERT-joint, and RoBERTa-joint.

\textbf{BiLSTM}, we use the same bidirectional LSTM (BiLSTM) to encode both the comment and reply, and the average hidden states of each word are regarded as sentence representations of them. The sentence representations of the comment and reply are then concatenated. We use a linear layer to reduce the dimension, after which a softmax layer is applied to obtain the label's probability distribution. We use Glove-300 as the initial word embedding, a popular word embedding method capturing semantics \cite{pennington2014glove}.

\textbf{BiLSTM-rel}, we concatenate the textual information encoded by BiLSTM with the relation feature $\textbf{h}^R$ and use a linear layer and a  softmax layer to identify the (dis)agreement. 

\textbf{BERT-joint}, we feed the input of $[CLS] \ comment \ [SEP]\ reply\ [SEP]$ into the BERT  and apply a linear layer to reduce the dimension of $[CLS]$ hidden states, after which a softmax layer is used to obtain the distributions.

\textbf{BERT-sep}, the comment and reply are encoded by BERT in the format of $[CLS] \ comment \ [SEP]$ and $[CLS] \ reply \ [SEP]$ separately. The hidden states of $[CLS]$ tokens are concatenated as the representations of the comment-reply pair for classification.

\textbf{RoBERTa-joint}, we feed the input of $[CLS] \ comment \ [SEP]\ reply $ $\ [SEP]$ into the RoBERTa and apply a linear layer to reduce the dimension of $[CLS]$ hidden states, after which a softmax layer is used to obtain the distributions.

\subsection{In-domain Results}
\subsubsection{Overall results}

We train our model with all the data from five subreddits, and the results are shown in Table \ref{results1}. First, BiLSTM achieves 47.56\%, 54.00\%, and 32.70\% macro-F1 scores for the categories, respectively, which are the lowest compared with BERT-based and RoBERTa-based models. It indicates that pre-trained language models such as BERT and RoBERTa can better learn textual representations for (dis)agreement detection. BiLSTM-rel achieves 53.75\%, 53.77\%, and 34.17\% macro-F1 scores for the classes, respectively. The macro-F1 score is 47.23\% for BiLSTM-rel, which is 2.48\% higher than that of BiLSTM. It demonstrates that the graph autoencoder and social relation information can help boost the performance of the randomly initialized model in the disagreement detection task.

\begin{table}[]

\begin{center}

\begin{tabular}{lcccccccccccc}

\hline      & r/Br & r/Cl  & r/BLM & r/Re &r/De\\

      \hline\hline
BiLSTM & 44.82  &43.08&51.81&46.59 &52.86       \\
BERT-joint   & 64.10 &64.90 &66.90 & {66.10} &{67.20}    \\
BERT-sep  &63.68 & 65.05&64.24&65.11&66.73\\
RoBERTa-joint &65.83&66.92&71.23&69.38&67.55\\
\hline
BiLSTM-rel &46.15&44.46&53.89&50.05&53.27\\
BERT-rel  &{65.99} &66.99&{70.17}&67.77&67.04\\
RoBERTa-rel &\textbf{66.81}&\textbf{68.77}&\textbf{71.37}&\textbf{70.25}&\textbf{68.24}\\
\hline

\end{tabular}
\caption{Accuracies of RoBERTa-rel on each subreddit.}
      \label{each}
      \vspace{-1cm}
\end{center}

\end{table}

\begin{table*}[]

\begin{center}

\begin{tabular}{lccccccccccccccc}

\hline   \multirow{2}{*}{Model}            & \multicolumn{2}{c}{r/Br} & \multicolumn{2}{c}{r/Cl}  & \multicolumn{2}{c}{r/BLM} & \multicolumn{2}{c}{r/Re} &\multicolumn{2}{c}{r/De} &\multicolumn{2}{c}{Average}\\ \cline{2-13} 
&Acc &M-F1 &Acc &M-F1 &Acc &M-F1 &Acc &M-F1&Acc &M-F1 &Acc &M-F1
\\
      \hline\hline
BiLSTM & 42.60  &41.90&41.52&40.24&46.11&39.73&47.30&41.32  &50.88&46.79&45.68 &42.00    \\
BERT-sep &61.84 &61.73  &63.82 & 63.11 &65.80 &62.86 &64.23&61.51& 65.91&63.52&64.32&62.71\\
BERT-joint   &{{64.12}} &{62.56}&64.42&64.34&65.32&62.13&{66.64}&63.25& 66.03 & 63.21 &65.30 &63.10 \\

RoBERTa-joint &65.43&64.07&67.64&65.95&69.15&66.06&66.02&64.94&64.80&61.45&66.61&64.46\\
\hline
\textbf{Ours} \\

BiLSTM-rel & 44.32 &43.19 & 42.33 & 43.14 &46.33 & 41.05&49.32&44.13&50.78&48.14&46.62&43.93\\

BERT-rel  &\textbf{66.49}&\textbf{65.13}&65.44&64.05&68.30&65.60&66.57&64.38&64.22&62.43 &66.20&64.32\\
RoBERTa-rel&{66.03}&64.49&\textbf{68.29}&\textbf{66.83} &\textbf{69.17}&\textbf{66.49}&\textbf{70.23}&\textbf{67.88}&\textbf{67.96}&\textbf{66.88}&\textbf{68.34}&\textbf{66.51}\\
\hline

\end{tabular}
\caption{Cross-domain testing results. The models are trained on the four subreddits and tested on the left subreddit.}
      \label{cross}
      \vspace{-0.7cm}
\end{center}

\end{table*}

In addition, the averaged macro-F1 model of BERT-joint is 0.32\% higher than that of BERT-sep, indicating that the joint model can perform better than the pipeline model for the former captures the attention features between comments and replies. Our model BERT-rel achieves 70.35\%, 72.34\%, and 53.51\% macro-F1 scores for the data, respectively. The averaged marco-F1 score of BERT-rel is 65.40\%, which is 2.13\% higher than that of BERT-joint and 1.89\% higher than that of BERT-sep, indicating that social relation information has a significant effect on a pre-trained (dis)agreement detection model. 

Moreover, RoBERTa-rel achieves the macro-F1 score of 66.91\%, which is a state-of-the-art performance, which is 1.24\% higher than that of RoBERTa-joint and 1.51\% higher than that of BERT-rel. The results show that the model can achieve stronger performance with more accurate textual information features, and social relation information can still enhance the (dis)agreement detection performance. RoBERTa-rel achieves macro-F1 scores of 71.44\%, 74.27\%, and 55.95\% on the labels, respectively. The improvement is the most significant on the data of disagreeing labels (2.38\% higher than RoBERTa-joint), where the relations of \textit{opponent} in the aggregate social graph are also more than those of \textit{supporter}. And the most challenging part of the dataset is still the neutral data due to the small proportion of the neutral data (statistics are in Table \ref{table1}).

\subsubsection{Breakdown results}
We test the models (trained on the five subreddits) on the data of each subreddit, and the results are shown in Table \ref{each}. The accuracies of BiLSTM-rel are 46.15\%, 44.46\%, 53.89\%, 50.05\%, and 53.27\% for the subreddits of Brexit, Climate, BlackLivesMatter, Republican, and Democrats, respectively, which are all higher than those of BiLSTM, indicating the effectiveness of social relation information. The same phenomenon can also be observed in the BERT-rel and RoBERTa-rel, indicating our social relation is effective for a different model architecture of (dis)agreement detection. RoBERTa-rel achieves the accuracies of 66.81\%, 68.77\%, 71.37\%, 70.25\%, and 68.24\% for the subreddits Brexit, Climate, BlackLivesMatter, Republican, and Democrats, respectively, which all achieve the state-of-the-art performance. The accuracy of RoBERTa-rel on BlackLivesMatter is improved the least (0.14\%), which results from the sparsity of the edges in the data of the BlackLivesMatter (2,516 nodes, 19.29 edges, and 1.51 averaged degree).

\begin{figure}
    \centering
    \includegraphics[width=0.88\hsize]{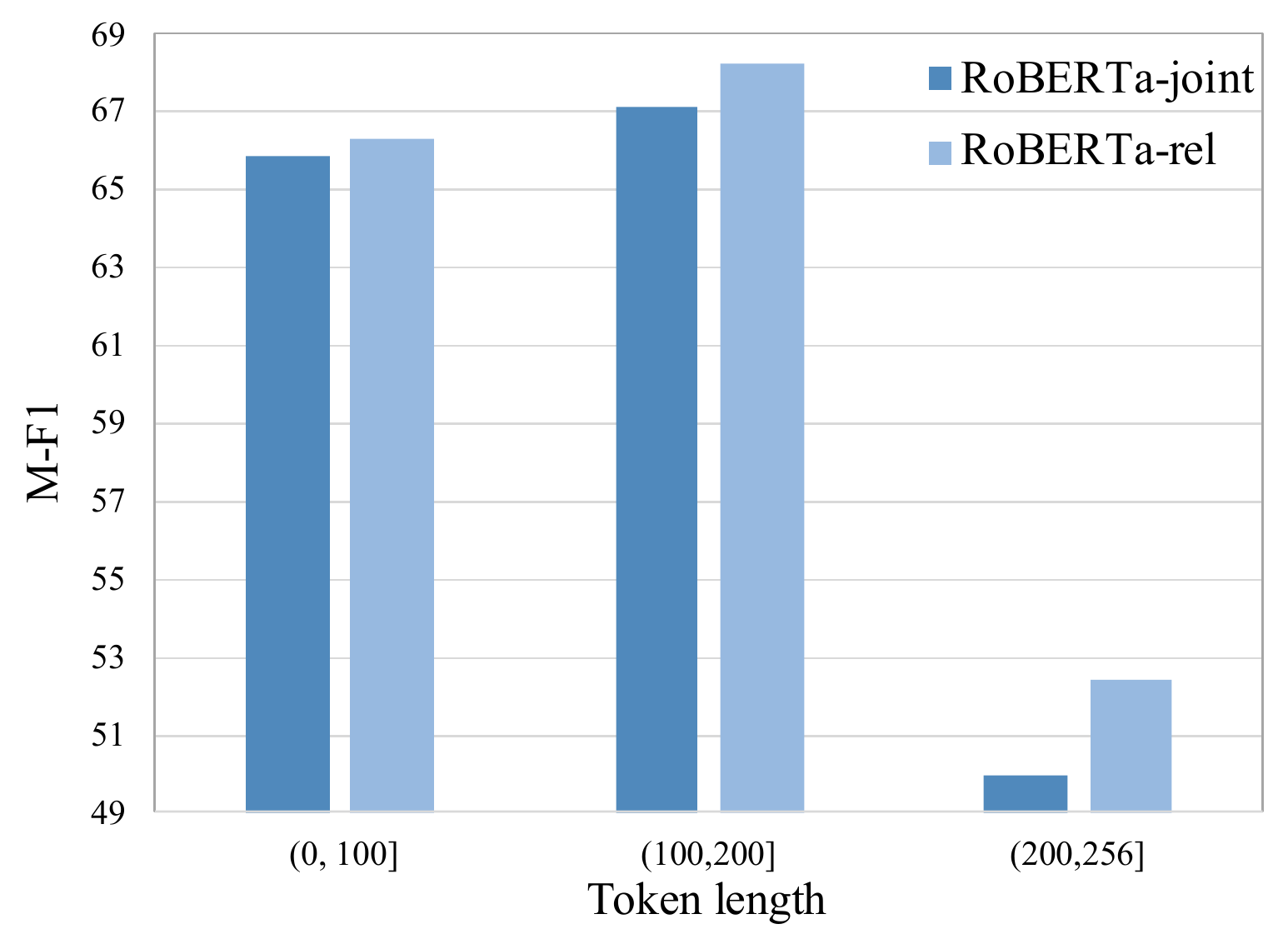}
    \vspace{-3mm}
    \caption{Results of RoBERTa-rel with respect to the different token lengths of the comment-reply pairs. }
    \label{length}
    \vspace{-0.5cm}
\end{figure}

\subsection{Cross-domain Results}
We evaluate our model in the cross-domain settings, which aims to evaluate the model generalization ability, reducing the cost and requirement of human annotations for models \cite{chen2022unisumm,bai2022cross,luo2022mere,zengjiali}. In particular, we train our model on the data of four sudreddits and test it on the left subreddit. The results are shown in Table \ref{cross}. The macro-F1 scores of BiLSTM are 41.90\%, 40.24, 39.73\%, 41.32\%, and 46.79\% on each task, which is the worst compared with BERT-based and RoBERTa-based models, indicating that the randomly initialized model are less informative in the features for the task of (dis)agreement detection. The model BiLSTM-rel achieves the 43.19\%, 43.14\%, 41.05\%, 44.13\%, and 48.14\% on each tasks, which are 1.29\%, 2.90\%, 0.32\%, 2.81\% and 1.45\% higher than those of BiLSTM, respectively. The results show that by using social relations, the model can achieve stronger performances.

As in-domain testing results, BERT-joint can still perform better than BERT-sep, but both are less effective than BERT-rel in the cross-domain settings. 
 The averaged precision and macro-F1 scores of BERT-rel on all the subreddit are 66.20\% and 64.32\%, which are 0.9\%, 1.22\% higher than BERT-joint, and 1.88\%, 1.61\% higher than BERT-sep, respectively. The results demonstrate the effectiveness of social relations in assisting (dis)agreement detection. Our model RoBERTa-rel achieves 68.34\% accuracy and 66.51\% macro F1 score on average, which is the best performance of our model on the (dis)agreement detection task. The performance is the lowest in the Brexit subreddit due to the large averaged degree and betweenness (35.39 and 1.54) in the subreddit while deleting the data from training hinders the model to learn complete social information (shown in Table \ref{graph}). But in other subreddit, the macro F1 scores show a roughly positive correlation with an averaged degree and betweenness of the subgraphs in each subreddit (i.e., with the increase of averaged degree and betweenness, the improvement margin of macro F1 score increases). In particular, the averaged degrees are 0.01, 0.22, 0.49, and 0.52 for the subgraphs in the subreddits BlackLivesMatter, Republican, Climate, and Democrats, and the corresponding improvement margins are 0.43\%, 3.05\%, 0.49\%, and  5.43\%, respectively. The phenomenon demonstrates that with more abundant social relation information, it is simpler to identify the (dis)agreement.   Note that the results of climate departure from the positive correlation, which may result from the reason that the authors of the Climate subreddit have less relation to those in other subreddits.
 
\vspace{-2.5mm}
 

\subsection{Further Analysis}

\subsubsection{Effect of token lengths.} We test our model  RoBERTa-rel with respect to different token lengths of comment-reply pairs (shown in Figure \ref{length}). It shows that RoBERTa-rel boosts the averaged macro-F1 scores of (dis)agreement detection with a large margin compared with RoBERTa-joint, 1.87\% and 2.45 for the data with lengths $(100,200]$ and $>200$, respectively, but it outperforms RoBERTa-joint only 0.45\%  for data with lengths $(0,100]$. The results show that it becomes challenging to identify the (dis)agreement labels with long sequence lengths merely using textual information. And it demonstrates that social relation information boosts the performance (dis)agreement detection, especially for the data with long lengths, which are difficult for models merely using textual information.

\subsubsection{Fusing Method.}
We also test our model with other methods for fusing textual and social relation information. We add the feature of social relation information and textual information, following $p= Softmax(W(\textbf{h}^{CLS}+\textbf{h}^{R})+b)$. The averaged macro F1 score and accuracy are 66.54\% and 67.86\%, which are 0.37\% and 0.52\% lower than those of concatenating method. It demonstrates that although concatenation is intuitive, it is more effective than addition.

\begin{figure}
    \centering
    \includegraphics[width=0.9\hsize]{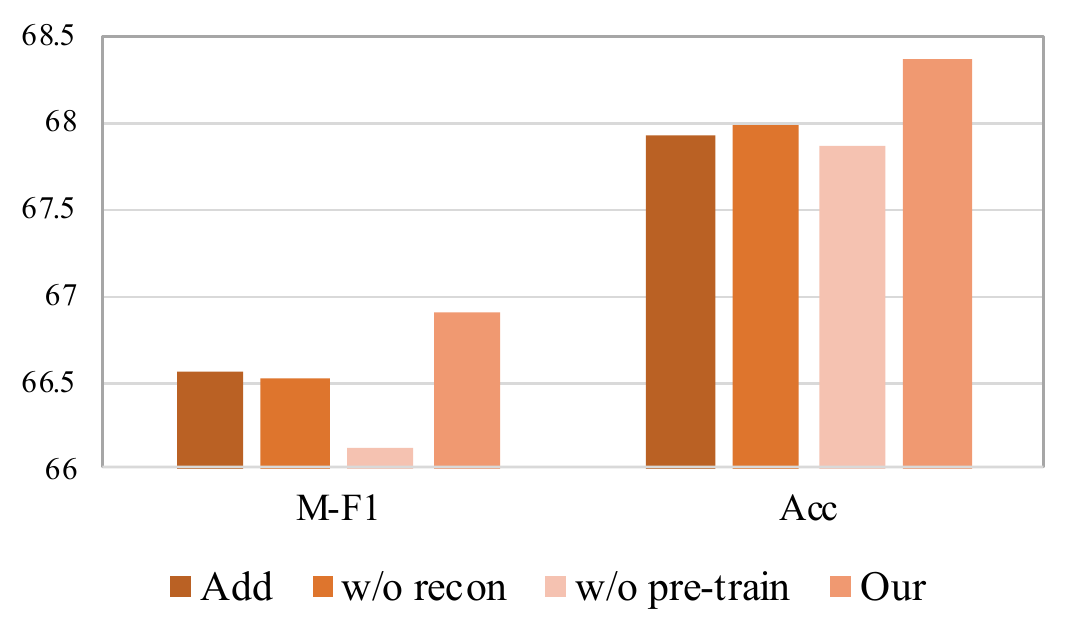}
    \vspace{-5mm}
    \caption{Ablation study on RoBERTa-rel, and different methods of information fusing in the in-domain testing. }
    \label{abl}
    \vspace{-4mm}
\end{figure}

 \begin{table*}[thp]
  \centering
\begin{tabular}{p{0.32\textwidth}p{0.32\textwidth}p{0.07\textwidth}p{0.05\textwidth}p{0.05\textwidth}}
\hline

{\textbf{Comment}} & \textbf{Reply}&\textbf{Soci Rel.}&\textbf{Label} &\textbf{Output} \\
\hline
\hline
By that standard, every person on the internet is hundreds of times more guilty than rural villagers in Africa and India. Why don't you give up your technology?  &  That wasn't the point. I just read a news article telling people what they can do to stop climate change when he himself has multiple private jets. He can take first class on a normal plane but that would inconvenience him. &Supporter&Agree &Agree\\ \hline

\hline
Am I the only person who gets worried when they see a line of only other people! lol. jokes but... actually not joking. It scares me now.  &  I smile (awkwardly, I'm sure) at poc. I'll knock a person up if anyone were to harass someone who's just minding their own business. &Interaction&Disagree&Disagree
\\
\hline
\end{tabular}
\caption{Case Study. Soci Rel. is for social relations.} 
\label{case}
\vspace{-7mm}
\end{table*}

\subsubsection{Ablation Study}
Figure \ref{abl} shows the results of ablation studies. First, we show the results without using the reconstruction loss function, but only cross-entropy loss for (dis)agreement classification. The averaged macro F1 score and accuracy are 66.25\% and 68.00\%, which are 0.39\% and 0.38\% lower than those of RoBERTa-rel, respectively.

We also test our model without pre-training the RGCN module using KGE methods but solely train it on the (dis)agreement objectives $\mathcal{L}_{train}$. Without pre-training the RGCN module, the model performance decreases with a large margin of 0.78\% and 0.80\% in averaged macro F1 score and accuracy, respectively. It demonstrates the significance of the pre-training process in the embeddings of the social relation graph.

\begin{table}[]

\begin{center}

\begin{tabular}{lcccc}

\hline      & Agree F1 & Disagree F1 & Neutral F1 & M-F1\\

      \hline\hline
RoBERTa-joint & 70.56&71.89&54.57&65.67    \\
TransE   &70.66&72.10&54.70&65.82\\
TransF  &71.12&72.22&54.75&66.03\\
HolE  &71.03 &73.34&54.92&66.43 \\
DistMult(Ours)  &\textbf{71.44}&\textbf{74.27}&\textbf{55.02}&\textbf{66.91}\\
\hline

\end{tabular}
\caption{Results with respect to different scoring functions of the graph autoencoder of the model ReBERTa-rel.}
      \label{score}
      \vspace{-1cm}
\end{center}

\end{table}

\subsubsection{Scoring Function of Graph Autoencoder}
To further analyze the influence of different knowledge graph embeddings (KGE), we compare  RoBERTa-rel (using the DistMult method) with several models using other typical scoring functions in the decoder of the graph autoencoder (the encoding method of the textual information is the same), including the translated-based methods \textbf{TransE} \cite{bordes2013translating}, \textbf{TransF} \cite{feng2016knowledge}, and semantic matching method \textbf{HolE} \cite{nickel2016holographic}. The results are shown in Table \ref{score}.

Observed in the results, all the models using the graph autoencoder outperform the model RoBERTa-joint, demonstrating the effectiveness of using social relation information.
The translational distance methods TransE (a macro-F1 score of 65.67\%) and TransF (a macro-F1 score of 65.82\%) perform worse than the semantic matching methods HolE (a macro-F1 score of 66.43\%) and DistMult (a macro-F1 score of 66.91\%), for the reason TransE and TransF only extract the relation information of entities instead of semantic information. Since we consider the social relation graph in a directed graph, HolE should obtain more substantial expressive power than DistMult in encoding asymmetric relations. However, the model with the HolE method achieves lower performance (66.43\%) than that with the DistMult method (66.91\%), which may result from the sparsity of the social relation graph.

\subsubsection{Effect of \textit{interaction} selection $\rho$}
\begin{figure}
    \flushleft
    \includegraphics[width=0.86\hsize]{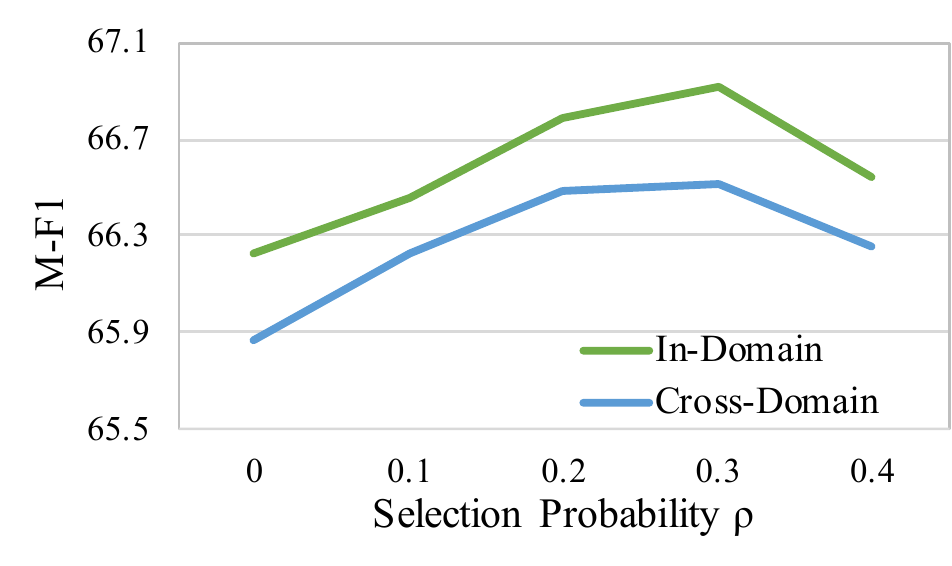}
    \vspace{-5mm}
    \caption{Results of RoBERTa-rel with respect to different rates of selected \textit{interaction} edges in the training set. For the in-domain task, the model is trained in all subreddits and evaluated on the test data. For the cross-domain task, the metric is the averaged macro-F1 score of the five tasks in Section 4.4.}
    \label{rho}
    \vspace{-5mm}
\end{figure}

We evaluate our model concerning different numbers of selected \textit{interaction} edges in the training set. The results are illustrated in Figure \ref{rho}. As is observed in the results, when $\rho$ is 0.0, the macro-F1 scores of both the in-domain and cross-domain tests are the lowest, which are 66.22\% and 65.77\%, respectively, which indicates the importance of adding \textit{interaction} edges with the increase of $\rho$. The macro-F1 scores increase and reach the peak (66.91\% and 66.51\%, respectively) when $\rho$ is 0.3, which means the selection of edges to be \textit{interaction} boost the performance of the (dis)agreement models.
However, as $\rho$ continues increasing to 0.4, the macro-F1 scores of both the tasks decrease (66.54\% and 66.25\%), which indicates that excessive \textit{interaction} relations can also introduce noise and spurious features to social relation information.

\subsection{Case Study}
Some cases are shown in Table \ref{case}.
The first case shows that the reply `That was not the point.' implies a disagreeing stance towards the comment, which results in incorrect identification of RoBERTa-joint. Benefiting from the social relation \textit{supporter} between them, RoBERTa-rel outputs a correct stance.
For the second case, for the complex textual information limited without textual context, BERT-joint fails to give the correct output.
Although there is only one \textit{interaction} relation between the authors of the comment ($n_c$) and the reply ($n_t$), the authors have the same neighbor in the relation graph who $n_c$ supporting but $n_t$ opposing. It implies the authors of this comment-reply pair may be opposing, and this information assists BERT-rel in outputting the correct stance. The cases show the effectiveness and reasonableness of using social relations.

\section{Conclusion}
We proposed a method to construct an inductive social relation graph from the comment-reply data to assist (dis)agreement detection. The model used a graph autoencoder to extract relation information, consisting of an RGCN encoder and a DistMult decoder for pre-training. Our model achieves state-of-the-art performance in the standard dataset DEBAGREEMENT for in-domain and cross-domain settings, showing social relations' effectiveness. We found social relation boosts the performance, especially for the long-token comment-reply pairs. Ablation studies showed the significance of each module. The study shows that general external information can boost the (dis)agreement detection. It is promising to model the opinions of the authors on different topics and further analyze how social relations form and how opinions spread on social platforms. For future work, it is a promising direction to consider leveraging the effective temporal information in the sparse social graph network, and in this way, it becomes 
feasible to study how public opinions spread and evolve on the social platform in more realistic settings.
 
\section*{Acknowledgement}
 We would  like to thank the anonymous reviewers for the detailed and thoughtful reviews. The work is funded by the  Pioneer and "Leading Goose" R\&D Program of Zhejiang under Grant Number 2022SDXHDX0003 and State Key Laboratory of Media Convergence Production Technology and System 2020 Annual Research Project (SKLMCPTS2020006).

\vspace{-1mm}
    

\bibliographystyle{ACM-Reference-Format}
\bibliography{sample-base}

\appendix

\begin{table*}[!t]
  \centering
\begin{tabular}{p{0.32\textwidth}p{0.32\textwidth}p{0.07\textwidth}p{0.05\textwidth}p{0.05\textwidth}}
\hline

{\textbf{Comment}} & \textbf{Reply}&\textbf{Soci Rel.}&\textbf{Label} &\textbf{Output} \\
\hline
\hline
Gates is promoting Exxon fantasy air carbon capture and "new" nuclear that is not in anyway close to being useful and would take way too long to build when we need cheap, fast and safe renewable energy to replace fossil fuels right now. He is promoting his own book and wants a position on Biden's climate team. &
 I read his book and in it he actually says that the air carbon capture in no way is scaleable enough, but whatever you say man. 
&Opponent&Disagree &Disagree\\ \hline

\hline
  What other countries are experiencing this? Needing to give up towns/big areas to water due to rising sea levels? &Greenland is ground zero for climate change, and everybody who lives in Greenland lives right on or very near the coast.
 &Interaction&Neutral&Neutral \\
 \hline 
 Think they'll do it? sounds like a cry for attention, surely they know this will render them politically incompetent. & This is the GOP splitting in 2 before our eyes. A lot of conservatives were horrified by the events on Jan. 6, and never bought the big lie. &Interaction&Agree&Agree \\ \hline
 A president isn't supposed to be impeached for failing to respond to the most pressing issues in your opinion. He should've been impeached very early on for breaking a handful of other guidelines of the presidency, abusing the power of the office, and violating the Constitution. His climate policy is not something impeachable.& That's absolutely ridiculous. His climate policy should be impeachable. Stupid rules and precedent aren't as important as preventing extinction. &Opponent & Disagree & Disagree  \\ \hline
\end{tabular}
\caption{Case Study. Soci Rel. is for social relations.} 
\label{case2}
\vspace{-7mm}
\end{table*}    

\section{Time period of the Construction of the Social Relations}
\begin{figure}[H]
    \centering
    \includegraphics[width=\hsize]{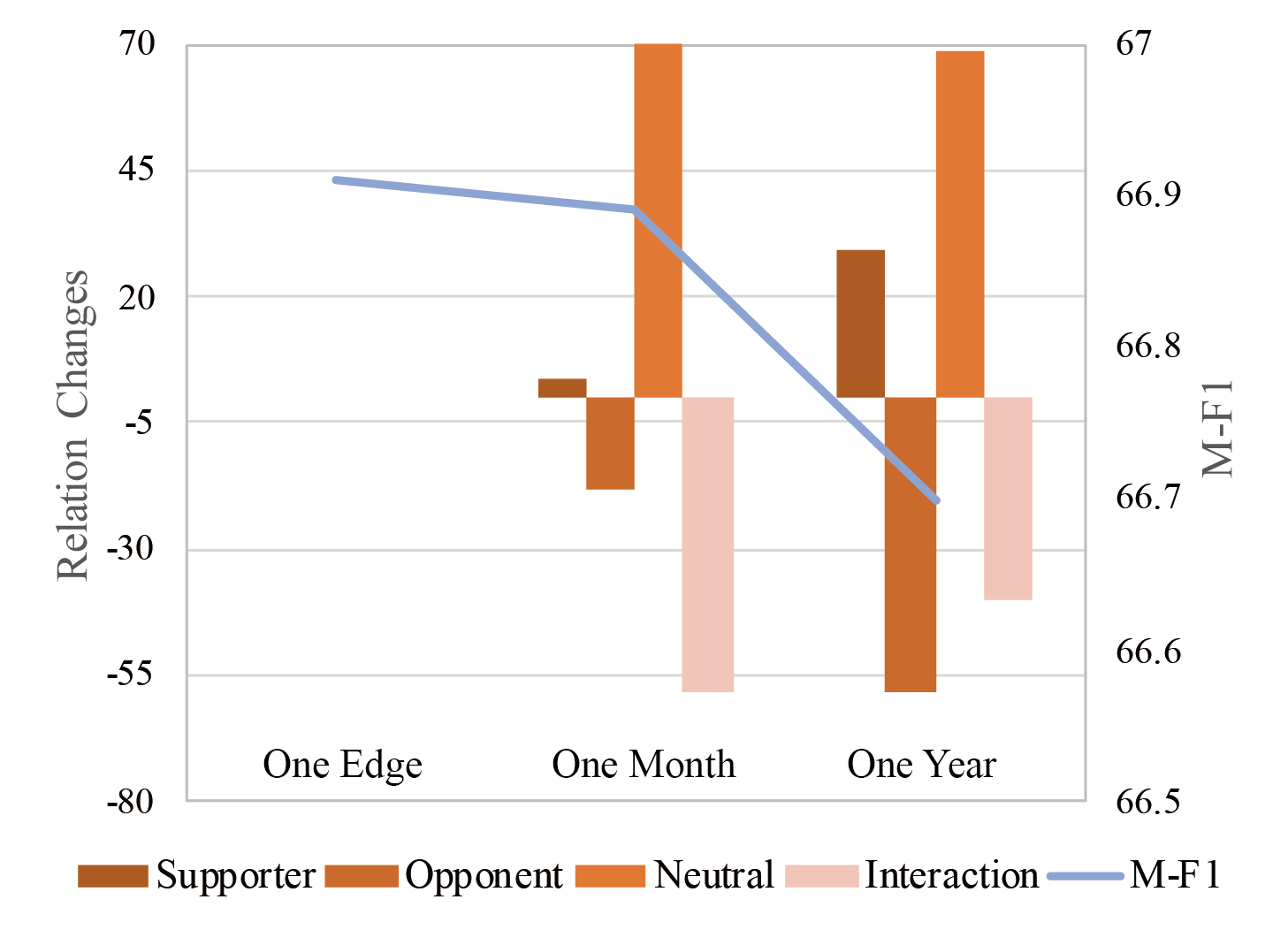}
    \caption{Results of RoBERTa-rel and the changes of relation numbers with respect to different time intervals.}
    \label{change}
\end{figure}

We also design the test with the time period $\tau$ in the construction of the social relation graph. The results are shown in Figure  \ref{change}. With the increase of the time period, the number of edges in different types varies in a small range (due to the sparsity of comment-reply pairs). We can still observe that with the increase of the time interval, the change of relations is more drastic, with the decrease of the model performance (from 66.91\%, 66.83\%, to 66.68\%). A suitable time interval is a significant part of the model due to the change in the effectiveness of inductive social relations. The results go against our common sense that the social relation has temporal effectiveness, while it may result from the sparsity of the comment-reply pairs. More deep analysis requires comprehensive datasets with multiple interactions between authors and dense graphs, which can be future work in such an area.

\section{Case Study}
We show more case studies in Table \ref{case2}.

\end{document}